# Future Frame Prediction for Anomaly Detection – A New Baseline

Wen Liu,[*] Weixin Luo,[*] Dongze Lian, Shenghua Gao[†]
ShanghaiTech University
{liuwen, luowx, liandz, gaoshh}@shanghaitech.edu.cn

## Abstract

*Anomaly detection in videos refers to the identification of events that do not conform to expected behavior. However, almost all existing methods tackle the problem by minimizing the reconstruction errors of training data, which cannot guarantee a larger reconstruction error for an abnormal event. In this paper, we propose to tackle the anomaly detection problem within a video prediction framework. To the best of our knowledge, this is the first work that leverages the difference between a predicted future frame and its ground truth to detect an abnormal event. To predict a future frame with higher quality for normal events, other than the commonly used appearance (spatial) constraints on intensity and gradient, we also introduce a motion (temporal) constraint in video prediction by enforcing the optical flow between predicted frames and ground truth frames to be consistent, and this is the first work that introduces a temporal constraint into the video prediction task. Such spatial and motion constraints facilitate the future frame prediction for normal events, and consequently facilitate to identify those abnormal events that do not conform the expectation. Extensive experiments on both a toy dataset and some publicly available datasets validate the effectiveness of our method in terms of robustness to the uncertainty in normal events and the sensitivity to abnormal events. All codes are released in https://github.com/StevenLiuWen/ano_pred_cvpr2018.*

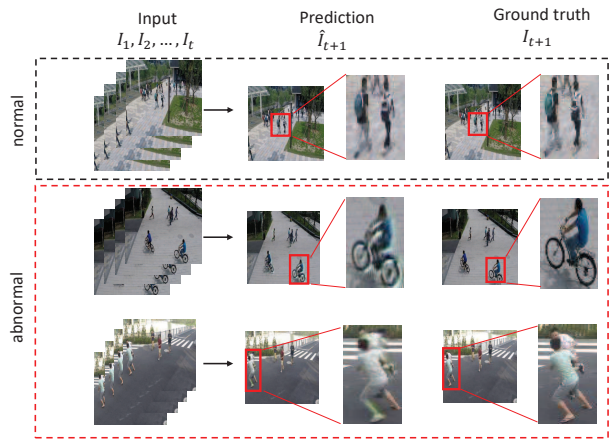

Figure 1. Some predicted frames and their ground truth in normal and abnormal events. Here the region is walking zone. When pedestrians are walking in the area, the frames can be well predicted. While for some abnormal events (a bicycle intrudes/ two men are fighting), the predictions are blurred and with color distortion. Best viewed in color.

## 1. Introduction

Anomaly detection in videos refers to the identification of events that do not conform to expected behavior [3]. It is an important task because of its applications in video surveillance. However, it is extremely challenging because abnormal events are unbounded in real applications, and it is almost infeasible to gather all kinds of abnormal events and tackle the problem with a classification method.

Lots of efforts have been made for anomaly detection [20][13][23]. Of all these work, the idea of feature reconstruction for normal training data is a commonly used strategy. Further, based on the features used, all existing methods can be roughly categorized into two categories: i) hand-crafted features based methods [6][20]. They represent each video with some hand-crafted features including appearance and motion ones. Then a dictionary is learnt to reconstruct normal events with small reconstruction errors. It is expected that the features corresponding to abnormal events would have larger reconstruction errors. But since the dictionary is not trained with abnormal events and it is usually overcomplete, we cannot guarantee the expectation. ii) deep learning based methods [13][5][26]. They usually learn a deep neural network with an Auto-Encoder way and they enforce it to reconstruct normal events with small reconstruction errors. But the capacity of deep neural network is high, and larger reconstruction errors for abnormal events do not necessarily happen. Thus, we can see that almost all training data reconstruction based methods cannot guaran-tee the finding of abnormal events.

---

[*]The authors contribute equally and are listed in alphabetical order.
[†]Corresponding author.

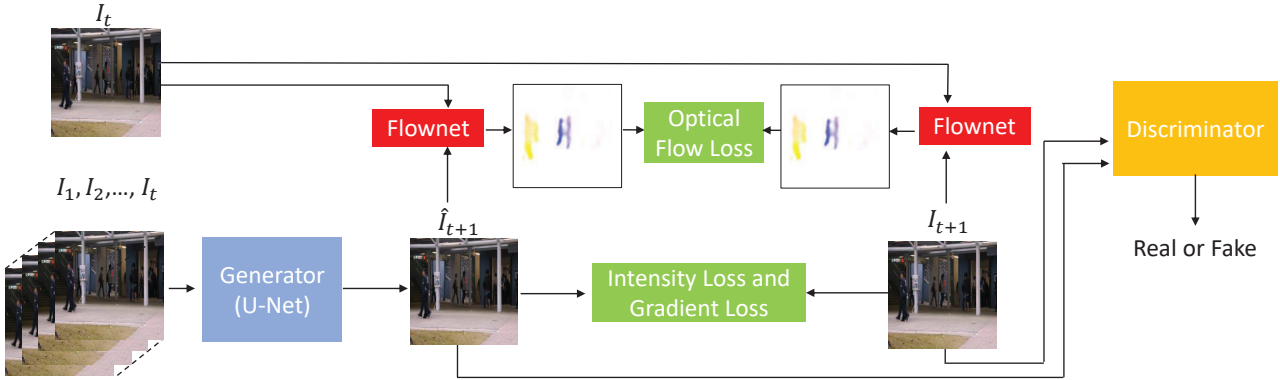

Figure 2. The pipeline of our video frame prediction network. Here we adopt U-Net as generator to predict next frame. To generate high quality image, we adopt the constraints in terms of appearance (intensity loss and gradient loss) and motion (optical flow loss). Here Flownet is a pretrained network used to calculate optical flow. We also leverage the adversarial training to discriminate whether the prediction is real or fake.

It is interesting that even though anomaly is defined as those events do not conform the expectation, most existing work in computer vision solve the problem within a framework of reconstructing training data [20][38][13]. We presume it is probable that the video frame prediction is far from satisfactory at that time. Recently, as the emergence of Generative Adversarial Network (GAN) [12], the performance of video prediction has been greatly advanced [25]. In this paper, rather than reconstructing training data for anomaly detection, we propose to identify abnormal events by comparing them with their expectation, and introduce a future video frame prediction based anomaly detection method. Specifically, given a video clip, we predict the future frame based on its historical observation. We first train a predictor that can well predict the future frame for normal training data. In the testing phase, if a frame agrees with its prediction, it potentially corresponds to a normal event. Otherwise, it potentially corresponds to an abnormal event. Thus a good predictor is a key to our task. We implement our predictor with an U-Net [28] network architecture given its good performance at image-to-image translation [15]. First, we impose a constraint on the appearance by enforcing the intensity and gradient maps of the predicted frame to be close to its ground truth; Then, motion is another important feature for video characterization, and a good prediction should be consistent with real object motion. Thus we propose to introduce a motion constraint by enforcing the optical flow between predicted frames to be close to their ground truth. Further, we also add a Generative Adversarial Network (GAN) [12] module into our framework in light of its success for video generation [25] and image generation [9].

We summarize our contributions as follows: i) We propose a future frame prediction based framework for anomaly detection. Our solution agrees with the concept of anomaly detection that normal events are predictable while abnormal ones are unpredictable. Thus our solution is more suitable for anomaly detection. To the best of our knowledge, it is the first work that leverages video prediction for anomaly detection; ii) For the video frame prediction framework, other than enforcing predicted frames to be close to their ground truth in spatial space, we also enforce the optical flow between predicted frames to be close to their optical flow ground truth. Such a temporal constraint is shown to be crucial for video frame prediction, and it is also the first work that leverages a motion constraint for anomaly detection; iii) Experiments on toy dataset validate the robustness to the uncertainty for normal events, which validates the robustness of our method. Further, extensive experiments on real datasets show that our method outperforms all existing methods.

## 2. Related Work

### 2.1. Hand-crafted Features Based Anomaly Detection

Hand-crafted features based anomaly detection is mainly comprised of three modules: i) extracting features; In this module, the features are either hand-crafted or learnt on training set; ii) learning a model to characterize the distribution of normal scenarios or encode regular patterns; iii) identifying the isolated clusters or outliers as anomalies. For feature extraction module, early work usually utilizes low-level trajectory features, a sequence of image coordinates, to represent the regular patterns [32][35]. However, these methods are not robust in complex or crowded scenes with multiple occlusions and shadows, because trajectory features are based on object tracking and it is very easy to fail in these cases. Taking consideration of the shortcom-ings of trajectory features, low-level spatial-temporal features, such as histogram of oriented gradients (HOG) [27], histogram of oriented flows (HOF) [7] are widely used. Based on spatial-temporal features, Zhang et al. [37] exploit a Markov random filed (MRF) for modeling the normal patterns. Adam et al. [2] characterize the regularly local histograms of optical flow by an exponential distribution. Kim and Grauman [16] model the local optical flow pattern with a mixture of probabilistic PCA (MPPCA). Mahadevan et al. [23] fit a Gaussian mixture model to mixture of dynamic textures (MDT). Besides these statistic models, sparse coding or dictionary learning is also a popular approach to encode the normal patterns [38][20][6]. The fundamental underlying assumption of these methods is that any regular pattern can be linearly represented as a linear combination of basis of a dictionary which encodes normal patterns on training set. Therefore, a pattern is considered as an anomaly if its reconstruction error is high and vice verse. However, optimizing the sparse coefficients is usually time-consuming in sparse reconstruction based methods. In order to accelerate both in training and testing phase, Lu et al [20] propose to discard the sparse constraint and learn multiple dictionaries to encode normal scale-invariant patches.

## 2.2. Deep Learning Based Anomaly Detection.

Deep learning approaches have demonstrated their successes in many computer vision tasks [18][11] as well as anomaly detection [13]. In the work [36], Xu *et al.* design a multi-layer auto-encoder for feature learning, which demonstrates the effectiveness of deep learning features. In another work [13], a 3D convolutional auto-encoder (Conv-AE) is proposed by Hasan to model regular frames. Further, motivated by the observation that Convolutional Neural Networks (CNN) has strong capability to learn spatial features, while Recurrent Neural Network (RNN)and its long short term memory (LSTM) variant have been widely used for sequential data modeling. Thus, by taking both advantages of CNN and RNN, [5][21] leverage a Convolutional LSTMs Auto-Encoder (ConvLSTM-AE) to model normal appearance and motion patterns at the same time, which further boosts the performance of the Conv-AE based solution. In [22], Luo *et al.* propose a temporally coherent sparse coding based method which can map to a stacked RNN framework. Besides, Ryota *et al.* [14] combine detection and recounting of abnormal events. However, all these anomaly detections are based on the reconstruction of regular training data, even though all these methods assume that abnormal events would correspond to larger reconstruction errors, due to the good capacity and generalization of deep neural network, this assumption does not necessarily hold. Therefore, reconstruction errors of normal and abnormal events will be similar, resulting in less discrimination.

## 2.3. Video Frame Prediction

Recently, prediction learning is attracting more and more researchers' attention in light of its potential applications in unsupervised feature learning for video representation [25]. In [29], Shi *et al.* propose to modify original LSTM with ConvLSTM and use it for precipitation forecasting. In [25], a multi-scale network with adversarial training is proposed to generate more natural future frames in videos. In [19], a predictive neural network is designed and each layer in the network also functions as making local predictions and only forwarding deviations. All aforementioned work focuses on how to directly predict future frames. Different from these work, recently, people propose to predict transformations needed for generating future frames [33] and [4], which further boosts the performance of video prediction.

## 3. Future Frame Prediction Based Anomaly Detection Method

Since anomaly detection is the identification of events that do not conform the expectation, it is more natural to predict future video frames based on previous video frames, and compare the prediction with its ground truth for anomaly detection. Thus we propose to leverage video prediction for anomaly detection. To generate a high quality video frame, most existing work [15][25] only considers appearance constraints by imposing intensity loss [25], gradient loss [25], or adversarial training loss [15]. However, only appearance constraints cannot guarantee to characterize the motion information well. Besides spatial information, temporal information is also an important feature of videos. So we propose to add an optical flow constraint into the objective function to guarantee the motion consistency for normal events in training set, which further boosts the performance for anomaly detection, as shown in the experiment section (section 4.5 and 4.6). It is worth noting abnormal events can be justified by either appearance (A giant monster appears in a shopping mall) or motion (A pickpocket walks away from an unlucky guy), and our future frame prediction solution leverages both the appearance and motion loss for normal events, therefore these abnormal events can be easily identified by comparing the prediction and ground truth. Thus the appearance and motion losses based video prediction are more consistent with anomaly detection.

Mathematically, given a video with consecutive $t$ frames $I_1, I_2, \ldots, I_t$, we sequentially stack all these frames and use them to predict a future frame $I_{t+1}$. We denote our prediction as $\hat{I}_{t+1}$. To make $\hat{I}_{t+1}$ close to $I_{t+1}$, we minimize their distance regarding intensity as well as gradient. To preserve the temporal coherence between neighboring frames, we enforce the optical flow between $I_{t+1}$ and $I_t$ and that between $\hat{I}_{t+1}$ and $I_t$ to be close. Finally, the difference between a future frame's prediction and itself determines whether it is normal or abnormal. The network architecture of our framework is shown in Fig. 2. Next, we will introduce all the components of our framework in details.

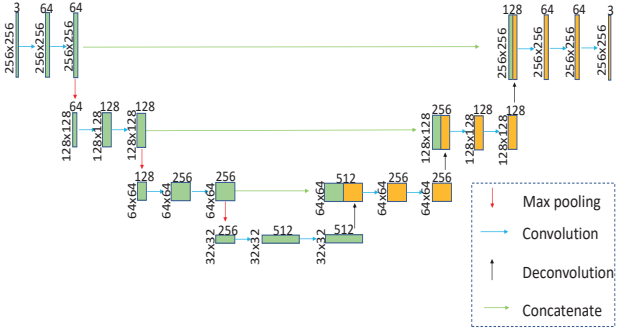

Figure 3. The network architecture of our main prediction network (U-Net). The resolutions of input and output are the same.

### 3.1. Future Frame Prediction

The network commonly used for frame generation or image generation in existing work [25][13] usually contains two modules: i) an encoder which extracts features by gradually reducing the spatial resolution; and ii) a decoder which gradually recovers the frame by increasing the spatial resolution. However, such a solution confronts with the gradient vanishing problem and information imbalance in each layer. To avoid this, U-Net[28] is proposed by adding a shortcut between a high level layer and a low level layer with the same resolution. Such a manner suppresses gradient vanishing and results in information symmetry. We slightly modify U-Net for future frame prediction in our implementation. Specifically, for each two convolution layers, we keep output resolution unchanged. Consequently, it does not need the crop and resize operations anymore when adding shortcuts. The details of this network are illustrated in Figure 3. The kernel sizes of all convolution and deconvolution are set to $3 \times 3$ and that of max pooling layers are set to $2 \times 2$.

### 3.2. The Constraints on Intensity and Gradient

To make the prediction close to its ground truth, following the work [25], intensity and gradient difference are used. The intensity penalty guarantees the similarity of all pixels in RGB space, and the gradient penalty can sharpen the generated images. Specifically, we minimize the $\ell_2$ distance between a predicted frame $\hat{I}$ and its ground true $I$ in intensity space as follows:

$$L_{int}(\hat{I}, I) = \|\hat{I} - I\|_2^2 \tag{1}$$

Further, we define the gradient loss by following previous work [25] as follows:

$$L_{gd}(\hat{I}, I) = \sum_{i,j} \Big\| |\hat{I}_{i,j} - \hat{I}_{i-1,j}| - |I_{i,j} - I_{i-1,j}| \Big\|_1 \\ + \Big\| |\hat{I}_{i,j} - \hat{I}_{i,j-1}| - |I_{i,j} - I_{i,j-1}| \Big\|_1 \tag{2}$$

where $i, j$ denote the spatial index of a video frame.

### 3.3. The Constraint on Motion

Previous work [25] only considers the difference between intensity and gradient for future frame generation, and it can not guarantee to predict a frame with the correct motion. This is because even a small change occurs in terms of the pixel intensity of all pixels in a predicted frame, even though it corresponds to a small prediction error in terms of gradient and intensity, it may result in totally different optical flow, which is a good estimator of motion [30]. So it is desirable to guarantee the correctness of motion prediction. Especially for anomaly detection, the coherence of motion is an important factor for the evaluation of normal events. Therefore, we introduce a temporal loss defined as the difference between optical flow of prediction frames and ground truth. However, the calculation of optical flow is not easy. Recently, a CNN based approach has been proposed for optical flow estimation [8]. Thus we use the Flownet [8] for optical flow estimation. We denote $f$ as the Flownet, then the loss in terms of optical flow can be expressed as follows:

$$L_{op} = \|f(\hat{I}_{t+1}, I_t) - f(I_{t+1}, I_t)\|_1 \tag{3}$$

In our implementation, $f$ is pre-trained on a synthesized dataset [8], and all the parameters in $f$ are fixed.

### 3.4. Adversarial Training

Generative adversarial networks (GAN) have demonstrated its usefulness for image and video generation [9][25]. By following [25], we also leverage a variant of GAN (Least Square GAN [24]) module for generating a more realistic frame. Usually GAN contains a discriminative network $\mathcal{D}$ and a generator network $\mathcal{G}$. $\mathcal{G}$ learns to generate frames that are hard to be classified by $\mathcal{D}$, while $\mathcal{D}$ aims to discriminate the frames generated by $\mathcal{G}$. Ideally, when $\mathcal{G}$ is well trained, $\mathcal{D}$ cannot predict better than chance. In practice, adversarial training is implemented with an alternative update manner. Moreover, we treat the U-Net based prediction network as $\mathcal{G}$. As for $\mathcal{D}$, we follow [15] and utilize a patch discriminator which means each output scalar of $\mathcal{D}$ corresponds a patch of an input image. Totally, the training schedule is illustrated as follows:

**Training $\mathcal{D}$.** The goal of training $\mathcal{D}$ is to classify $I_{t+1}$ into class 1 and $\mathcal{G}(I_1, I_2, ..., I_t) = \hat{I}_{t+1}$ into class 0, where 0 and 1 represent fake and genuine labels, respectively. When training $\mathcal{D}$, we fix the weights of $\mathcal{G}$, and a Mean Square Error (MSE) loss function is imposed:

$$L_{adv}^{\mathcal{D}}(\hat{I}, I) = \sum_{i,j} \frac{1}{2} L_{MSE}(\mathcal{D}(I)_{i,j}, 1) \\ + \sum_{i,j} \frac{1}{2} L_{MSE}(\mathcal{D}(\hat{I})_{i,j}, 0) \quad (4)$$

where $i, j$ denotes the spatial patches indexes and $L_{MSE}$ is a MSE function, which is defined as follows:

$$L_{MSE}(\hat{Y}, Y) = (\hat{Y} - Y)^2 \quad (5)$$

where $Y$ takes values in $\{0,1\}$ and $\hat{Y} \in [0, 1]$

**Training $\mathcal{G}$.** The goal of training $\mathcal{G}$ is to generate frames where $\mathcal{D}$ classify them into class 1. When training $\mathcal{G}$, the weights of $\mathcal{D}$ are fixed. Again, a MSE function is imposed as follows:

$$L_{adv}^{\mathcal{G}}(\hat{I}) = \sum_{i,j} \frac{1}{2} L_{MSE}(\mathcal{D}(\hat{I})_{i,j}, 1) \quad (6)$$

### 3.5. Objective Function

We combine all these constraints regarding appearance, motion, and adversarial training, into our objective function, and arrive at the following objective function:

$$\begin{aligned} L_{\mathcal{G}} =& \lambda_{int} L_{int}(\hat{I}_{t+1}, I_{t+1}) \\ &+ \lambda_{gd} L_{gd}(\hat{I}_{t+1}, I_{t+1}) \\ &+ \lambda_{op} L_{op} \\ &+ \lambda_{adv} L_{adv}^{\mathcal{G}}(\hat{I}_{t+1}) \end{aligned} \quad (7)$$

When we train $\mathcal{D}$, we use the following loss function:

$$L_{\mathcal{D}} = L_{adv}^{\mathcal{D}}(\hat{I}_{t+1}, I_{t+1}) \quad (8)$$

To train the network, the intensity of pixels in all frames are normalized to [-1, 1] and the size of each frame is resized to $256 \times 256$. We set $t = 4$ and use a random clip of 5 sequential frames which is the same with [25]. Adam [17] based Stochastic Gradient Descent method is used for parameter optimization. The mini-batch size is 4. For gray scale datasets, the learning rate of generator and discriminator are set to 0.0001 and 0.00001, respectively. While for color scale datasets, the learning rate of generator and discriminator start from 0.0002 and 0.00002, respectively. For different datasets, the coefficient factors of $\lambda_{int}, \lambda_{gd}, \lambda_{op}$ and $\lambda_{adv}$ are slightly different. An easy way is to set $\lambda_{int}, \lambda_{gd}, \lambda_{op}$ and $\lambda_{adv}$ as 1.0, 1.0, 2.0 and 0.05, respectively.

### 3.6. Anomaly Detection on Testing Data

We assume that normal events can be well predicted. Therefore, we can use the difference between predicted frame $\hat{I}$ and its ground truth $I$ for anomaly prediction. MSE is one popular way to measure the quality of predicted images by computing a Euclidean distance between the prediction and its ground truth of all pixels in RGB color space. However, Mathieu [25] shows that Peak Signal to Noise Ratio (PSNR) is a better way for image quality assessment, shown as following:

$$PSNR(I, \hat{I}) = 10 \log_{10} \frac{[\max_{\hat{I}}]^2}{\frac{1}{N} \sum_{i=0}^{N} (I_i - \hat{I}_i)^2}$$

High PSNR of the $t$-th frame indicates that it is more likely to be normal. After calculating each frame's PSNR of each testing video, following the work [25], we normalize PSNR of all frames in each testing video to the range [0, 1] and calculate the regular score for each frame by using the following equation:

$$S(t) = \frac{PSNR(I_t, \hat{I}_t) - \min_t PSNR(I_t, \hat{I}_t)}{\max_t PSNR(I_t, \hat{I}_t) - \min_t PSNR(I_t, \hat{I}_t)}$$

Therefore, we can predict whether a frame is normal or abnormal based its score $S(t)$. One can set a threshold to distinguish regular or irregular frames.

## 4. Experiments

In this section, we evaluate our proposed method as well as the functionalities of different components on three publicly available anomaly detection datasets, including the CUHK Avenue dataset [20], the UCSD Pedestrian dataset [23] and the ShanghaiTech dataset [22]. We further use a toy dataset to validate the robustness of our method, i.e., even if there exists some uncertainties in normal events, our method can still correctly classify normal and abnormal events.

### 4.1. Datasets

Here we briefly introduce the datasets used in our experiments. Some samples are shown in Fig. 4.

- CUHK Avenue dataset contains 16 training videos and 21 testing ones with a total of 47 abnormal events, including throwing objects, loitering and running. The size of people may change because of the camera position and angle.

- The UCSD dataset contains two parts: The UCSD Pedestrian 1 (Ped1) dataset and the UCSD Pedestrian 2 (Ped2) dataset. The UCSD Pedestrian 1 (Ped1) dataset includes 34 training videos and 36 testing ones with

Table 1. AUC of different methods on the Avenue, Ped1, Ped2 and ShanghaiTech datasets.

|  | CUHK Avenue | UCSD Ped1 | UCSD Ped2 | ShanghaiTech |
|---|---|---|---|---|
| MPPCA [16] | N/A | 59.0% | 69.3% | N/A |
| MPPC+SFA [23] | N/A | 66.8% | 61.3% | N/A |
| MDT [23] | N/A | 81.8% | 82.9% | N/A |
| Conv-AE [13] | 80.0% | 75.0% | 85.0% | 60.9% |
| Del *et al.* [10] | 78.3% | N/A | N/A | N/A |
| ConvLSTM-AE [21] | 77.0% | 75.5% | 88.1% | N/A |
| Unmasking [31] | 80.6% | 68.4% | 82.2% | N/A |
| Hinami *et al.*[14] | N/A | N/A | 92.2% | N/A |
| Stacked RNN [22] | 81.7% | N/A | 92.2% | 68.0% |
| **Our proposed method** | **84.9%** | **83.1%** | **95.4%** | **72.8%** |

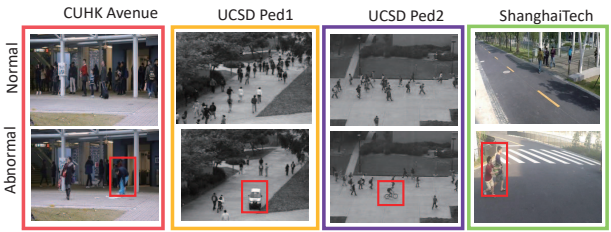

Figure 4. Some samples including normal and abnormal frames in the UCSD, CUHK Avenue and ShanghaiTech datasets are illustrated. Red boxes denote anomalies in abnormal frames.

40 irregular events. All of these abnormal cases are about vehicles such as bicycles and cars. The UCSD Pedestrian 2 (Ped2) dataset contains 16 training videos and 12 testing videos with 12 abnormal events. The definition of anomaly for Ped2 is the same with Ped1. Usually different methods are evaluated on these two parts separately.

- The ShanghaiTech dataset is a very challenging anomaly detection dataset. It contains 330 training videos and 107 testing ones with 130 abnormal events. Totally, it consists of 13 scenes and various anomaly types. Following the setting used in [22], we train the model on all scenes.

### 4.2. Evaluation Metric

In the literature of anomaly detection [20][23], a popular evaluation metric is to calculate the Receiver Operation Characteristic (ROC) by gradually changing the threshold of regular scores. Then the Area Under Curve (AUC) is cumulated to a scalar for performance evaluation. A higher value indicates better anomaly detection performance. In this paper, following the work [22], we leverage frame-level AUC for performance evaluation.

### 4.3. Comparison with Existing Methods

In this section, we compare our method with different hand-craft features based method [16][23][34][10] and latest deep learning based methods [13][31][14][22]. The AUC of different methods is listed in Table 1. We can see that our method outperforms all existing methods (around (3–5)% on all datasets), which demonstrates the effectiveness of our method.

### 4.4. The Design of Prediction Network

In our anomaly detection framework, the future frame prediction network is an important module. To evaluate how different prediction networks affect the performance of anomaly detection, we compare our U-Net prediction network with Beyond Mean Square Error (Beyond-MSE) [25] which achieves state-of-the-art performance for video generation. Beyond-MSE leverages a multi-scale prediction network to gradually generate video frames with larger spatial resolution. Because of its multi-scale strategy, it is much slower than U-Net. To be consistent with Beyond-MSE, we adapt our network architecture by removing the motion constraint and only use the intensity loss, the gradient loss and adversarial training in our U-Net based solution.

**Quantitative comparison for anomaly detection.** We first compute the gap between average score of normal frames and that of abnormal frames, denoted as $\Delta_s$. We compare the result of U-Net with that of Beyond-MSE on the Ped1 and Ped2 datasets, respectively. Larger $\Delta_s$ means the network can be more capable to distinguish normal and abnormal patterns. Then, we also compare the U-Net based solution and Beyond-MSE with the AUC metric on the Ped1 and Ped2 datasets, respectively. We demonstrate the results in Table 2. We can see that our method both achieves a larger $\Delta_s$ and higher AUC than Beyond-MSE, which show that our network is more suitable for anomaly detection than Beyond-MSE. Therefore, we adapt U-Net architecture as our prediction network. As we aforementioned, the results listed here do not contain motion constraint, which would further boost the AUC.

Table 2. The gap ($\Delta_s$) and AUC of different prediction networks in the Ped1 and Ped2 datasets.

|  | Ped1 | | Ped2 | |
| --- | --- | --- | --- | --- |
|  | $\Delta_s$ | AUC | $\Delta_s$ | AUC |
| Beyond-MSE | 0.200 | 75.8% | 0.396 | 88.5% |
| U-Net | **0.243** | **81.8%** | **0.435** | **93.5%** |

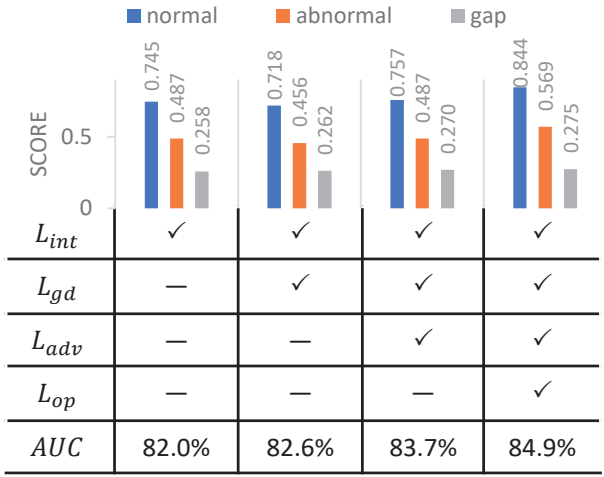

Figure 5. The evaluation of different components in our future frame prediction network in the Avenue dataset. Each column in the histogram corresponds to a method with different loss functions. We calculate the average scores of normal and abnormal events in the testing set. The gap is calculated by subtracting the abnormal score from the normal one.

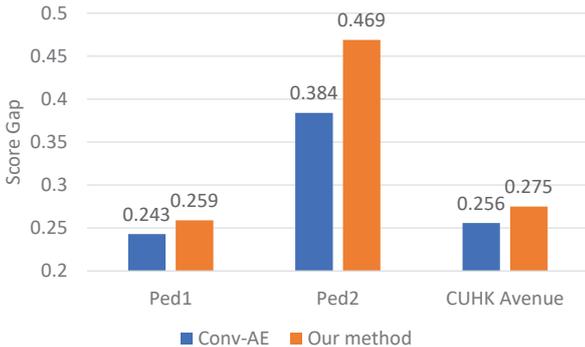

Figure 6. We firstly compute the average score for normal frames and that for abnormal frames in the testing set of the Ped1, Ped2 and Avenue datasets. Then, we calculate the difference of these two scores($\Delta_s$) to measure the ability of our method and Conv-AE to discriminate normal and abnormal frames. A larger gap($\Delta_s$) corresponds to small false alarm rate and higher detection rate. The results show that our method consistently outperforms Conv-AE in term of the score gap between normal and abnormal events.

Table 3. AUC for anomaly detection of networks with/wo the motion constraint in Ped1 and Ped2.

|  | Ped1 | Ped2 |
| --- | --- | --- |
| without motion constraint | 81.8% | 93.5% |
| with motion constraint | **83.1%** | **95.4%** |

### 4.5. Impact of Constraint on Motion.

To evaluate the importance of motion constraint for video frame generation as well as anomaly detection, we conduct the experiment by removing the constraint from the objective in the training. Then we compare such a baseline with our method.

**Evaluation of motion constraint with optical flow maps.** We show the optical flow maps generated with/without motion constraint in Fig. 7, we can see that the optical flow generated with motion constraint is more consistent with ground truth, which shows that such motion constraint term helps our prediction network to capture motion information more precisely. We also compare the MSE between optical flow maps generated with/without motion constraint and the ground truth, which is 7.51 and 8.26, respectively. This further shows the effectiveness of motion constraint.

**Quantitatively evaluation of motion with anomaly detection.** The result in Table 3 shows that the model trained with motion constraint consistently achieves higher AUC than that without the constraint on Ped1 and Ped2 dataset. This also proves that it is necessary to explicitly impose the motion consistency constraint into the objective for anomaly detection.

### 4.6. Impact of Different Losses for Anomaly Detection.

We also analyze the impact of different loss functions for anomaly detection by ablating different terms gradually. We combine different losses to conduct experiments on the Avenue dataset. To evaluate how different losses affect the performance of anomaly detection, we also utilize the score gap($\Delta_s$) mentioned above. The larger gap represents the more discriminations between normal and abnormal frames. The results in Figure 5 show more constraints usually achieve a higher gap as well as AUC value, and our method achieves the highest value under all settings.

### 4.7. Comparison of Prediction Network and Auto-Encoder Networks for Anomaly Detection

We also compare the video prediction network based and Auto-Encoder network based anomaly detection. Here for Auto-Encoder network based anomaly detection, we use the Conv-AE [13] which is the latest work and achieves state-of-the-art performance for anomaly detection. Because of the capacity of deep neural network, Auto-Encoder based methods may well reconstruct normal and abnormal frames in the testing phase. To evaluate the performance of prediction network and Auto-Encoder one, we also utilize the aforementioned gap($\Delta_s$) between normal and abnormal scores. The result in Fig. 6 shows that our solution always achieves higher gaps than Conv-AE, which validates the effectiveness of video prediction for anomaly detection.

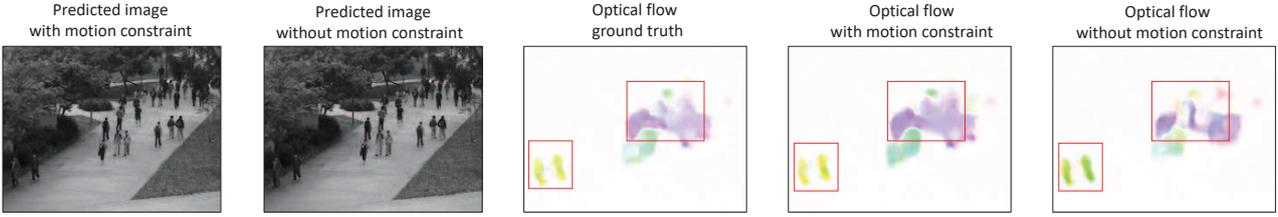

Figure 7. The visualization of optical flow and the predicted images on the Ped1 dataset. The red boxes represent the difference of optical flow predicted by the model with/without motion constraint. We can see that the optical flow predicted by the model with motion constraint is closer to ground truth. Best viewed in color.

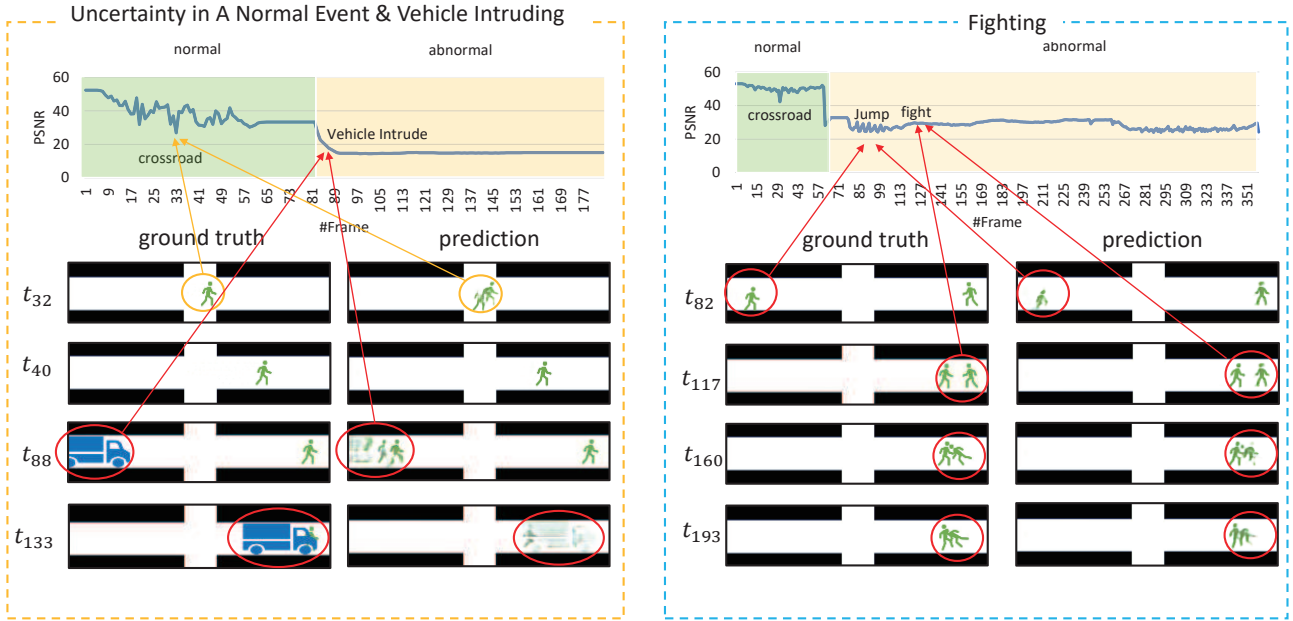

Figure 8. The visualization of predicted testing frames in our toy pedestrian dataset. There are two abnormal cases including vehicle intruding(left column) and humans fighting(right column). The orange circles correspond to normal events with uncertainty in prediction while the red ones correspond to abnormal events. It is noticeable that the predicted truck is blurred, because no vehicles appear in the training set. Further, in the fighting case, two persons cannot be predicted well because fighting motion never appear in the training phase.

### 4.8. Evaluation with A Toy Dataset

We also design a toy pedestrian dataset for performance evaluation. In the training set, only a pedestrian walks on the road and he/she can choose different directions when he/she comes to a crossroad. In the testing set, there are some abnormal cases such as vehicles intruding, humans fighting, *etc.*. We have uploaded our toy dataset in the supplementary material. Totally, the training data contains 210 frames and testing data contains 1242 frames.

It is interesting that the motion direction is sometimes also uncertain for normal events, for example, a pedestrian stands at the crossroad. Even though we cannot predict the motion well, we only cannot predict the next frame at a moment which leads a slightly instant drop in terms of PSNR. After observing the pedestrian for a while when the pedestrian has made his or her choice, it becomes predictable and PSNR would go up, shown in Fig. 8. Therefore the uncertainty of normal events does not affect our solution too much. However, for the real abnormal events, for example, a truck breaks into the scene and hits the pedestrian, it would leads to a continuous lower PSNR, which facilitates the anomaly prediction. Totally, the AUC is 98.9%.

## 4.9. Running Time

Our framework is implemented with NVIDIA GeForce TITAN GPUs and Tensorflow [1]. The average running time is about 25 fps, which contains both the video frame generation and anomaly prediction. We also report the running time of other methods such as 20 fps in [31], 150 fps [20] and 0.5 fps in [38].

## 5. Conclusion

Since normal events are predictable while abnormal events do not conform to the expectation, therefore we propose a future frame prediction network for anomaly detection. Specifically, we use a U-Net as our basic prediction network. To generate a more realistic future frame, other than adversarial training and constraints in appearance, we also impose a loss in temporal space to ensure the optical flow of predicted frames to be consistent with ground truth. In this way, we can guarantee to generate the normal events in terms of both appearance and motion, and the events with larger difference between prediction and ground truth would be classified as anomalies. Extensive experiments on three datasets show our method outperforms existing methods by a large margin, which proves the effectiveness of our method for anomaly detection.